\documentclass[journal,transmag]{IEEEtran}
\usepackage{graphicx}
\usepackage{graphics}
\usepackage{subfigure}
\usepackage{calligra}

\usepackage{mathalfa}
\usepackage{amsmath}
\usepackage{amsfonts}
\usepackage{amssymb}
\usepackage{amsmath}
\usepackage{fancyhdr}

\ifCLASSINFOpdf
\else
\fi
\begin{document}
%
\title{Understanding over-parameterized deep networks by geometrization}



\author{\IEEEauthorblockN{Xiao Dong, Ling Zhou}
\IEEEauthorblockA{Faculty of Computer Science and Engineering, Southeast University, Nanjing, China}}

%



\IEEEtitleabstractindextext{%
\begin{abstract}
A complete understanding of the widely used over-parameterized deep networks is a key step for AI. In this work we try to give a geometric picture of over-parameterized deep networks using our geometrization scheme. We show that the Riemannian geometry of network complexity plays a key role in understanding the basic properties of over-parameterizaed deep networks, including the generalization, convergence and parameter sensitivity. We also point out deep networks share lots of similarities with quantum computation systems. This can be regarded as a strong support of our proposal that geometrization is not only the bible for physics, it is also the key idea to understand deep learning systems.
\end{abstract}

\begin{IEEEkeywords}
over-parameterization, deep networks, geometrization, physics, quantum computation, Riemannian geometry
\end{IEEEkeywords}}

\maketitle



\IEEEdisplaynontitleabstractindextext

%
\IEEEpeerreviewmaketitle

\section{Motivation}

\emph{Are all layers created equal?} is a recent work which addressed the problem of how sensitive are the parameters in an over-parameterized deep network\cite{Bengio2019layer}. Their experiments show a heterogeneous characteristics of layers, where bottom layers have a higher sensitivity than top layers. This is an exciting observation since this is \emph{exactly} what the geometry of quantum computation told us about deep networks one decade ago!

In our former work\cite{Dong2019geo}, inspired by the facts that deep networks are effective descriptors for our physical world and deep networks share similar geometric structures of physical systems such as geometric mechanics, quantum computation, quantum many-body systems and even general relativity, we proposed a geometrization scheme to interpret deep networks and deep learning systems. The observation of \cite{Bengio2019layer} encouraged us to apply this scheme on over-parameterized deep networks to give a geometric description of such networks.

In the following parts of this paper, we will explore the similarities between deep networks and quantum computation systems. We will transfer the rich geometric structure of quantum mechanics and quantum computation systems to deep networks so that we have an intuitive geometric understanding of the basic properties of over-parameterized deep networks, including network complexity, generalization, convergence and the geometry formed by deep networks.

\section{Geometrization}
Geometrization of physics is the greatest and the most successful idea in understanding the rules of our physical world in human history. But why can our world be geometrized? In the last decade, we saw a new trend to combine geometrization and quantum information processing to draw a complete new picture of our world. Basically this is to regard our world, including spacetime, material and the interactions among them, as emergent from a complex quantum deep network. From this point of view, our world is built from deep networks and the geometric structure of the physical world emerges from the geometric structure of the underlying deep networks. So the geometrization of physics \emph{is} essentially the geometrization of the underlying quantum deep networks. The success of geometrization of physics indicates that geometrization is also the key to understand deep networks.

The similarities between deep networks and physical systems, including both classical geometric mechanics and quantum computation systems, have been addressed in our former works\cite{Dong_deep}\cite{Dong2019geo}. Here for simplicity we only give a brief recap of key points we have learned from the geometrization of quantum information processing that will be involved in this paper.

\subsection{Geometry of quantum information processing}
It's well known that quantum mechanics has a rich geometric structure so that we believe quantum mechanics is the ultimate rule of our world. Quantum information processing or quantum computation, which explores the complex structure of both quantum states and quantum state evolutions, is the ultimate tool to describe our world and the rules of quantum information processing systems can be applied to all physical systems, including deep networks. So what do we know already about quantum information processing systems?

\textbf{\emph{Gigantic quantum state space and the corner of physical states}} For simplicity we use the most popular model of quantum information processing, i.e. a quantum state is described by a n-qubit system and the quantum information processing is described by a quantum circuit model. The quantum state space is huge since the dimension of a n-qubit pure state system is $2^n$ and the number of possible states is $O(2^{2^n})$. In all the $O(2^{2^n})$ states, only a tiny zero measure subset, the corner of physical states, is physically realizable since the states in this subset can be generated with a polynomial complexity from a simple initial state such as the product state $|00...0\rangle$.

\textbf{\emph{Quantum computational complexity}} The concept of quantum computational complexity plays a key role not only in quantum computation but also in quantum gravity, black hole information problem and quantum phase transition\cite{Ge2018Quantum}\cite{Heydari_dynamicdiatance}\cite{Matsueda2013Emergent}\cite{Nielsen_geometry2}\cite{Susskind2016The}\cite{Susskind_ER_bridge}\cite{Susskind_ER_bridge_nowhere}. Basically a quantum algorithm on a n-qubit system is an unitary transformation $U\in \mathbf{U}(2^n)$ and its computational complexity $C(U)$ is given by the geodesic distance between the identity operation $I$ and $U$, where the geodesic is defined on the Riemannian manifold of $\mathbf{U}(2^n)$. For more details on the geometry of quantum computation, please refer to \cite{Nielsen_geometry}\cite{Nielsen_geometry2}. Accordingly the state complexity of a n-qubit quantum system $|\psi\rangle$ is defined as the minimal complexity of all the quantum algorithms that can  generate $|\psi\rangle$ from $|00...0\rangle$, i.e. $C(|\psi\rangle)=\min(C(U), |\psi\rangle=U|00...0\rangle)$. Since the DOF of a general n-qubit transformation $U \in \mathbf{U}(2^n)$ is $O(2^n)$, obviously its computational complexity is $O(2^n)$. This is to say, a general n-qubit algorithm can only be achieved by a quantum circuit with $O(2^n)$ quantum gates, which is regarded as non-realizable. What we are interested are the polynomial complexity algorithms, which can be used to prepare the corner of physical states from the product state $|00...0\rangle$.

\textbf{\emph{Quantum computational complexity and geometry}} Quantum computational complexity has a rich geometrical structure. Firstly the quantum complexity is defined on the Riemannian structure of the manifold $\mathbf{U}(2^n)$. A natural question is then, what's the curvature of the Riemannian manifold of quantum computation? It's shown that this manifold may have a non-positive curvature everywhere\cite{Nielsen_geometry}\cite{Nielsen_geometry2}. This is to say, the geodesic on this manifold is not stable and it's initial momentum sensitive. Keen readers can immediately see that we have now a connection between quantum computation and the observation of \cite{Bengio2019layer}. Secondly, the concept of quantum computational complexity builds a correspondence or a duality between quantum states and quantum algorithms. That's to say, given a quantum state $|\psi\rangle$, we have a correspondent optimal quantum algorithm $U(|\psi\rangle)$ to prepare it from an initial product state. If we take the quantum circuit of the algorithm $U(|\psi\rangle)$ as a network of quantum operations, then we have a duality between quantum states and quantum deep networks. This duality may play a key role in understanding the geometry of spacetime\cite{Swingle2009Entanglement}\cite{Swingle2012Constructing}\cite{Dong2018GR}. In fact the geometry of spacetime is just the geometry of the quantum deep network. The take-home message is, \emph{the dual quantum deep network of a quantum state is determined by a Riemannian geometry of the quantum transformation space, and a quantum deep network also generates a Riemannian geometry}. So do we have two Riemannian structures? There are signs to show, if we use the Fisher-Rao metric of the deep network, then they can be united and general relativity can be deduced from it\cite{Matsueda2014Derivation}\cite{Dong2019geo}.

\textbf{\emph{Quantum mechanics and geometry}} Finally, even we consider the most classical quantum mechanics without the fancy concept of quantum complexity, we can also learn something that can be applied to understanding deep networks. The first observation is the geometry of quantum state space. It's well known that quantum mechanics show a probabilistic property so that in a projective measurement, the probability that the state falls in an eigen state of the observable is determined by the distance between the initial state and the final state. Geometrically this means the probabilistic property of quantum mechanics is determined by the Riemannian structure of quantum mechanics. The second observation is the geometry of quantum evolution. A general quantum state evolution of a n-qubit system can be written as a sequence of unitary transformations $U_nU_{n-1}...U_1$ with $U_i \in \mathbf{U}(2^n)$. Obviously this can be regarded as a linear deep network. How about the stability of this system? It has been shown that this system show a chaotic property, which means a tiny perturbation of the first operation $U_1$ will lead to a huge change of the composite operation $U_nU_{n-1}...U_1$.

We will see all the afero-mentioned observations can help us to understand over-parameterized deep networks.

\section{Geometrization of over-parameterized deep networks}

\subsection{Over-parameterized deep networks}
We first give a brief summary of the known facts and arguments about over-parameterized deep networks.

\textbf{\emph{Over-parameterization}}
By over-parameterized deep networks, we usually mean the number of network parameters is much larger than the number of training data. The over-parameterization is in both the width and the depth of deep networks. Existing works show that over-parameterization plays a key role in the network capacity, convergency, generalization and even the acceleration of the optimization. But how exactly the over-parameterization can affect the performance of deep networks remains not completely clear to us.

\textbf{\emph{Local minima and convergence}}
It's obvious that over-parameterized networks have a large number of local minima. In \cite{Soudry2016No} it's shown that for over-parameterized deep network, with a high probability, all the local minima are also global minima as far as the data are not degenerated. A similar argument in \cite{Du2018Gradient}\cite{Allen2018A} told us that for sufficiently over-parameterized deep networks, gradient descent can reach local minima with a high probability from any initialization point of the network. Of course this is because the over-parameterization re-shaped the loss landscape of deep networks. Can we have an intuitive geometric picture of this point?

\textbf{\emph{Network complexity and generalization}}
Although all the local minima can all fit the training data well, we know they are not equal since they have different generalization capabilities and we prefer to find out a configuration with good generalization performance. Generally the generalization of a network is related with the network complexity\cite{Liang2017Fisher} and a lower network complexity means a better generalization performance. In \cite{Lei2017Towards} it's shown that the minima that can generalize well have a larger volume of basin of attraction so that they dominate over the poor ones. This is an interesting observation and we will show this is essentially an analogue of the probabilistic characteristics of quantum mechanics and it has a geometrical origin.

\textbf{\emph{Loss landscape}}
Over-parameterization changes the loss landscape.  \cite{Cooper2018Landscape} claimed that the locus of global minima is usually not discrete but rather an continuous high-dimensional submanifold of the parameter space. But how the structure of this submanifold changes with the number of parameters is still an open problem.

\textbf{\emph{Implicit acceleration by over-parameterization}}
In \cite{Arora2018acceleration} it's claimed that over-parameterization, especially in the depth direction, works as an acceleration mechanism for the optimization of deep networks and also this acceleration can not be achieved by a regularization. We will show maybe this is a misunderstanding of the role of over-parameterization.

\textbf{\emph{Layers are not created equal}}
For a multilayer deep network, it's a direct question to check if all the layers are equal. The recent work \cite{Bengio2019layer} showed that layers have different sentivities for either fully connected networks, convolutional networks or residual networks. What's the geometry behind this observation? We will try to understand this point as an analogue of quantum information processing systems.

\subsection{Geometric picture of over-parameterized deep networks}
The geometrization of deep networks has been explained in \cite{Dong_deep}\cite{Dong2019geo}, where we showed that deep networks share the same geometric structure of geometric mechanics and quantum computation systems. The key observation is that deep networks are curves to connect the identity transformation and the target transformation on the Riemannian manifold of data transformations. We will now see how over-parameterized deep networks can be understood in this geometrization framework.

\textbf{\emph{Over-parameterization}}
What's the role of over-parameterization in deep network? How to determine if a network is properly over-parameterized? In fact we can understand over-parameterization by comparing it with quantum computation systems. In quantum computation we have a gigantic state space and only a zero measure subset, the corner of physical states, is physically realizable. The duality between quantum states and quantum algorithms shows that this is also true for quantum algorithms. Similarly the space of possible functions between the input and output data of deep networks is also huge and only a small subset of it is physically interesting for us, which is the subset of functions that have a polynomial computational complexity. So essentially approximating a function by deep networks is to explore this subset. Compared with quantum computation systems, an universal shallow network is just a general unitary transformation $U \in U(2^n)$, which needs an exponential complexity to describe a transformation of data state space. A polynomial deep network is just a polynomial quantum circuit that only generate the corner of physical states. From this complexity point of view, deep networks are not really \emph{universal} since they only explore a subset of all possible transformations. In over-parameterized deep networks, increasing the width and depth of the networks can be understood as increasing the number of qubits and the length of the quantum circuit to achieve a quantum algorithm. A key point is that, in order to achieve a quantum algorithm $U$ the complexity of the quantum circuit, which is roughly proportional to the depth of the quantum circuit, has to exceed the quantum complexity of $U$.

\textbf{\emph{Local minima and convergence}}
How the over-parameterization can change the distribution of local minima and convergence is not very clear yet. If we compare deep networks with quantum mechanics, we can only say the cost function of deep networks can be regarded as a frustration free Hamiltonian and the global minima are ground states of the frustration free Hamiltonian. This observation is closely related with the concepts of parent Hamiltonian and uncle Hamiltonian. But if there is an exact correspondence between them is still under investigation.

\textbf{\emph{Network complexity and generalization}}
The relationship between network complexity and generalization capability is straight forward. In our former work to compare deep networks with the image registration problem, we indicated that the network complexity can be understood as the deformation energy of a diffeomorphic image transformation. So a lower network complexity means a smooth low energy deformation. Obviously a smooth image transformation has a better generalization performance. The observation of \cite{Lei2017Towards} that a solutions with a better generalization has a higher probability to be found during optimization from a random initialization then has an exact correspondence in quantum mechanics. As mentioned in the first section, during a projective measurement, the probability of a final quantum state $|\psi_f\rangle$ appears is related with its distance to the initial quantum state $|\psi_i\rangle$. This is to say, the probability $p(|\psi_i\rangle,|\psi_f\rangle)$ is determined by the complexity $C(U(|\psi_i\rangle,|\psi_f\rangle))$ of quantum transformation $U$ that transform the initial state to the final state so that $p(|\psi_i\rangle,|\psi_f\rangle)\sim e^{-C(U(|\psi_i\rangle,|\psi_f\rangle))}$. We see this is exactly what happens in over-parameterized deep networks. Here a better generalizaiton means a lower network complexity and a higher probability that this network configuration is found during optimization. Obviously we also have a relationship $p\sim e^{-C}$ between the probability and the complexity. So we can claim that the probability that a deep network configuration is found by optimization is determined by the network complexity, which is geometrically the Riemannian distance  between the transformation achieved by this network and the identity transformation $I$. It's very interesting to see classical deep networks show the same probabilistic property of quantum mechanics. For us it's more interesting to check if this observation can be used to understand quantum mechanics from a deep network point of view, because the measurement problem of quantum mechanics is still not fully understood. Can the commonly used decoherence picture of quantum measurement can be formulated as a training process of deep networks?

\textbf{\emph{Loss landscape}}
It's straight forward to see that over-parameterized deep network has a locus of global minima as an high-dimensional submanifold of the parameter space.
But we are not clear about the exact structure of this submanifold and how it will change with the increasing number of network parameters. For example, we have no idea if this high-dimensional submanifold is a connected or a separated manifold or even has a fractal-like structure. We highly suspect that the locus of global minima has a fractal structure since the network is nonlinear and the sensitivities of different layers are different as will be further addressed in the following discussions.

\textbf{\emph{Implicit acceleration by over-parameterization}}
Can the over-parameterization provide an implicit acceleration of the optimization as claimed in \cite{Arora2018acceleration}? To clarify this, we first restate the argument of \cite{Arora2018acceleration}, in which a linear neural network is considered as follows:
$\mathcal{X}:=\mathbb{R}^d$ and $\mathcal{Y}:=\mathbb{R}^k$ are the input and output data space. A $N$-layer linear network is used to fit a training set ${(x_i,y_i)}_{i=1}^m\in \mathcal{X} \times \mathcal{Y}$ and the $\mathcal{l}_2$ loss function $\sum_{i=1}^m(\hat{y}_i-y_i)^2$ is used, where $\hat{y}_i$ is the output of the network given the input $x_i$. The parameters of the depth-$N$ linear network are ${W_1,W_2,...,W_N}$ and the end-to-end weight matrix is given by $W_e=W_NW_{n-1}...W_1$ so that $L^N(W1,W2,...,W_N)=L^1(W_e)$. The gradient descent based optimization of $W_e$ can then be written as

\begin{equation}\label{eq-0}
\begin{split}
   W_e^{(k+1)}\Leftarrow & (1-\eta\lambda N)W_e^{(k)}-\eta\sum_{j=1}^N[W_e^{(k)}(W_e^{(k)})^\top]^{\frac{j-1}{N}}\\
   \cdot &\frac{dL^1(W_e^{(k)})}{dW}\cdot[(W_e^{(k)})^\top W_e^{(k)}]^{\frac{N-j}{N}}
\end{split}
\end{equation}

where they assume $W_{j+1}^{\top}(k)W_{j+1}(k)=W_{j}(k)W_{j}(k)^{\top},j=1,2,...,N-1$ is fulfilled for the network. \cite{Arora2018acceleration} argued that the difference between the N-layer deep network and a 1-layer network is that the gradient $\frac{dL^1(W_e^{(t)})}{dW}$ is transformed by the two items $[W_e^{(k)}(W_e^{(k)})^\top]^{\frac{j-1}{N}}$ and $[W_e^{(k)}(W_e^{(k)})^\top]^{\frac{N-j}{N}}$. They interpreted the effect of overparameterization (replacing clasic linear model by dept-N linear networks) on gradient descent as the deep network structure reshapes the gradient $\frac{dL^1(W_e^{(t)})}{dW}$ by changing both its amplitude and direction so that this can be understood as introducing some forms of momentum and adaptive learning rate. Also they claimed that this over-parameterization effect can not be obtained by regularization.

Do we have a geometric description of this observation in our geometrization scheme? In fact this can be directly observed by comparing deep networks with diffeomorphic image registration problem as in \cite{Dong_deep}\cite{Dong2019geo}. What's more, we can directly generalize the conclusion of \cite{Arora2018acceleration} to a general nonlinear deep network without any further assumptions on the network.

Diffeomorphic image registration can be abstracted as a map $G\times V\rightarrow V$, where $G$ is the group of image transformations and $V$ is the vector space of images. Large deformation diffeomorphic metric mapping (LDDMM)\cite{Beg2004Computing} generates a deformation $\varphi$ as a flow $\varphi^u_t$ of a time-dependent vector field $u_t \in T_e(G)=\mathbf{g}$ so that
\begin{equation}\label{eq1}
  \dot{\varphi}^u_t=u_t\circ\varphi^u_{t}, \varphi^u_{0}=Id, \varphi^u_{1}=\varphi
\end{equation}

The diffeomorphic matching of two images $I_0$ and $I_1$ with LDDMM is to find a vector field $u_t, t\in[0,1]$ to minimize the cost function
\begin{equation}\label{eq2}
\begin{split}
  E(u_t)=&E_K(u_t)+E_C(u_t)\\
        =&\int_0^1\frac{1}{2}|u_t|^2dt+\beta|I_1-I_o\circ\varphi^u_{1}|^2
\end{split}
\end{equation}
Here the regularity on $u_t$ is a kinetic energy term $E_K(u_t)=\frac{1}{2}\int_0^1|u_t|^2dt$ with $|u_t|$ a norm on the vector field defined as $|u_t|^2=\langle Lu_t,u_t\rangle_{L^2}$. The operator $L$ is a positive self-adjoint differential operator. Obviously the norm $|u_t|^2=\langle Lu_t,u_t\rangle_{L^2}$ defines a Riemannian metric on the manifold of the diffeomorphic transformation group $Diff(R^n)$. The second term $E_C(u_t)=\beta|I_1-I_o\circ\varphi^u_{1}|^2$ is the difference between the transformed image $I_o\circ\varphi^u_{1}$ and the target image $I_1$.

A necessary condition $DE(u_t)=0$ to minimize the cost function is that the vector field $u_t$ should satisfy the Euler-Poincar\'{e} (E-P) equation
\begin{equation}\label{eq3}
  Lu_t=-\varphi^u_{0,t}I_0\diamond \varphi^u_{0,t}\varphi^u_{1,0}\pi
\end{equation}
where $\varphi^u_{s,t}=\varphi^u_t\circ\varphi^u_{s^{-1}}$, $\pi:=\beta(\varphi^u_{0,t}I_0-I1)^\flat \in V^*$. The $\flat$ operator is defined as $\flat:V\rightarrow V^*,\langle u^{\flat},v\rangle_{V^*\times V}=\langle u,v\rangle$ and $\diamond:TV^*\rightarrow \mathbf{g}^*,\langle I\diamond \pi,u\rangle_{\mathbf{g}^*\times \mathbf{g}}=\langle\pi,\zeta_u(I)\rangle_{V^*\times V}$ is the momentum map.

In LDDMM framework, the curve satisfying the E-P equation is found by a gradient descent algorithm, while the gradient is given by $u_t+K\varphi^u_{0,t}I_0\diamond \varphi^u_{0,t}\varphi^u_{1,0}\pi$ with $K=L^{-1}$. A direct calculation in the LDDMM framework following \cite{Beg2004Computing} shows that the update of $\varphi^u_1$ is given by

\begin{equation}\label{eq-4}
\begin{split}
   \varphi^{u,(k+1)}_1 \Leftarrow & (1-\eta)\varphi^{u,(k)}_1\\
                                 -&\eta K\star\int_0^1 D\varphi^u_{t,1}\cdot D\varphi^u_{t,1} \cdot \frac{dE_C(u_t)}{d\varphi_1}\varphi^u_{0,t}\varphi^u_{0,t}dt
\end{split}
\end{equation}

We can directly see this is almost the same as the update rule of $W_e$ given by (\ref{eq-0}). But here we are working with a nonlinear deep network so that we have a generalization of the linear network of \cite{Arora2018acceleration}. In fact the result of \cite{Arora2018acceleration} can be regarded as a special case of LDDMM called static vector flow (SVF), which is formulated on a Lie group instead of on a Riemannian manifold and the items $[W_e^{(k)}(W_e^{(k)})^\top]^{\frac{j-1}{N}}$, $[W_e^{(k)}(W_e^{(k)})^\top]^{\frac{N-j}{N}}$ can be understood as an analogue of the Lie exponential used in SVF framework.

LDDMM has a beautiful geometric picture which is the same as the geometric mechanics\cite{Bruveris2011The}\cite{Bruveris2013Geometry}\cite{Holm2009Geometric}. How to understand the effect of over-parameterization in this LDDMM framework? LDDMM formulates a smooth image transformation by a constrained curve described by (\ref{eq1}). The gradient descent based update of the curve is essentially a constrained optimal control as shown in \cite{Hart2013An}. So when we try to approximate a function by deep networks, the structure of over-parameterized deep network is essentially to set constraints on the possible solution space. The so-called acceleration effect of over-parameterization in \cite{Arora2018acceleration} is nothing but a natural result of the constrained optimal control formulation. Also their conclusion that this acceleration can not be obtained by regularization is also not exact since the constraints in optimal control can also be regarded as a kind of regularization in optimization problems\cite{Holm2009Euler}. The only difference is that the regularization is set on the structure of the network.

\textbf{\emph{Layers are not created equal}}
We have seen that in quantum computation, for both the general sequential unitary quantum evolution and the quantum circuit model, we observe the same initial value sensitivity property. This is to say, quantum information processing systems are playing with Riemannian manifolds with negative curvatures. If we compare these with the observation of \cite{Bengio2019layer}, we find the general quantum evolution system corresponds to the fully connect networks and the quantum circuit model corresponds to convolutional networks. So we can say the observed non-equality of layers in \cite{Bengio2019layer} is just a direct consequence of the principle of quantum computation system. But there is still one thing is missing, the residual network. It's observed in \cite{Bengio2019layer} that residual networks also show a non-equality of layers but the pattern is different from fully connected and convolutional networks. Can we also find the correspondence of residual networks in quantum computation systems? Yes, since residual networks are just differential equations, they are correspondent to the fundamental quantum mechanics rule, the Schrodinger equation. Since the finite time discretization of Schrodinger equation is just the general sequential unitary quantum evolution, we believe Schrodinger equation should have the same initial value sensitivity pattern. This means residual networks should have a similar pattern as the fully connected and convolutional networks. This is different from the observed pattern of residual networks\cite{Bengio2019layer}. How to resolve this contradiction? If we believe that quantum mechanics is the ultimate rule of the world and the main advantage of residual networks is to build a smoother manifold of transformations to approximate functions, then residual networks should be related with a smooth geometry and there is no reason that some layers of residual networks are more critical than other layers as observed in \cite{Bengio2019layer}. We assume this is due to the artifacts of the non-uniform discretization used in residual networks and noise during optimization. From another aspect, the redistribution of the sensitivity pattern of residual networks also indicates that the strong background negative curvature geometry of general deep networks is weakened in residual networks so that the random perturbation effects survive. This is in fact an evidence that residual networks are building and working on a flatter manifold than fully connected and convolutional networks.

Another problem is related with the spacetime structure. There is evidence that the geometry of spacetime is emergent from quantum information processing networks. Also in \cite{Dong2019geo} we indicated that in deep networks, if the Fisher-Rao metric is used to measure the network complexity, then the interaction between data and network structures is analogue of the interaction between material and spacetime geometry, i.e. the general relativity. But if a general quantum deep network has a negative curvature, how can our universe have a flat (in a large scale) spacetime? Does the existence of our flat universe is an evidence that there exists a subset of deep networks that can form a flat Euclidean geometry? If such a corner of Euclidean deep networks exist, then all the layers will be created equal in such networks. Can this help us to find better network structures? In random matrix based analysis of deep networks, a special type of network configuration with dynamic isometry property seems to fall in this subset. It has been shown that such kind of networks hold some advantages beyond normal networks such as a smooth information flow in both the forward and backward directions. In fact geometrically the smooth information flow is just the inertial movement in a flat spacetime, i.e. the first law of Newton. Of course, just as the corner of physical states in quantum mechanics, the corner of Euclidean deep networks is also a zero measurement set. So we assume this subset may not form an universal data processing system, just as our universe may be a very special case of the so-called multiverse picture.

Finally, the negative curvature will influence the loss landscape of deep networks. If a network configuration has a higher sensitivity at the bottom layers, it can be easily figured out that loss landscape is more sensitive to the bottom layers and more robust to top layers. Accordingly the locus of the global minima will have more valleys in the bottom layers and the locus may have a fractal-like complex pattern with a stronger over-parameterization. How exactly the over-parameterization will change the loss landscape is still open.

\section{Conclusions}
Geometrization is not only the key idea of physics, it's also a framework to understand deep networks. In this work we try to understand over-parameterized deep networks by geometrization. By establishing analogies between properties of over-parameterized deep networks and quantum information processing/diffeomorphic image registration systems, we found they share similar geometric structures. Our key observations are:(1)Polynomial complexity over-parameterized deep networks only explore a corner of polynomial complexity functions just as quantum computation systems only explore the corner of physical states in the gigantic quantum state space. The network structure sets constraints on the submanifold of functions that can be approximated by the network. (2)Over-parameterized deep networks may have a complex loss landscape and local minima have different generalization capabilities. The generalization capability is determined by the network complexity, which is computed as the geodesic distance on a Riemannian manifold between the transformation represented by the network and the identity transformation. The probability that a certain configuration is obtained is determined by the complexity of the network. This is an analogue of the measurement problem in quantum mechanics, where the probability of the final state is determined by the distance between the initial state and the final state. (3)Over-parameterized deep networks have a geometry with a negative curvature, just as quantum computation systems has a Riemannian geometry with a negative curvature. All these observations suggest that deep networks are closely related with physics and geometrization may provide a proper roadmap to interpret deep networks.

In this work we mainly explore the Riemannian structure of deep networks, for example the network complexity as the geodesic distance and the sensitivity of network parameters as Riemannian curvature. A natural question is, can other geometrical structures in physics help to understand over-parameterized deep networks? For example the symplectic structure of geometric mechanics plays a key role in the dynamics of classical mechanics. Can the dynamics of deep networks also be understood in a similar way? Fibre bundle structure is another key structure to understand interactions in physics, also it plays a key role in the geometry of quantum information processing such as the geometry of mixed state and quantum entanglement. Can it be used to understand interactions between subnetworks in a composite system with multiple subnetworks? In \cite{Dong2019geo} we have mentioned that fibre bundles may be related with important network structures such as attention mechanism, Turing neural machines and differential neural computers. There are signs that fibre bundles are also related with capsule networks and the recent quaternion neural networks. To explore the possibility to understand deep networks based on bundles will be our future work.

\bibliographystyle{unsrt}

\bibliography{overparameterization}





\end{document}